\documentclass[conference]{IEEEtran}
\IEEEoverridecommandlockouts
\usepackage{amsmath,amsfonts}
\usepackage{algorithmic}
\usepackage{algorithm}
\usepackage{array}
\usepackage{textcomp}
\usepackage{stfloats}
\usepackage{url}
\usepackage{verbatim}
\usepackage{graphicx}
\usepackage{cite}
\usepackage{multirow}
\usepackage{graphicx} 
\usepackage{subcaption}
\usepackage{booktabs}

\begin{document}

\title{Deep Reinforcement Learning for Advanced Longitudinal Control and Collision Avoidance in High-Risk Driving Scenarios\\

\thanks{This work is supported  NDSU VPR Office project, Accelerating the Deployment of Autonomous Vehicles in Rural Areas, and partially supported by the National Science Foundation Award \#2234292 OAC Core: Stochastic Simulation Platform for Assessing Safety Performance of Autonomous Vehicles in Winter Seasons.}
\thanks{ $\IEEEauthorrefmark{1}$ D. Chen, Y. Gong, and XT. Yang are with the College of Engineering, University of Maryland, College Park, MD 20742, USA. (Email: \{dwchen98, ybgong, xtyang\}@umd.edu).
\textit{(Corresponding author: Xianfeng Terry Yang.)}}
}

\author{
		\IEEEauthorblockN{Dianwei Chen$\IEEEauthorrefmark{1}$,
		Yaobang Gong$\IEEEauthorrefmark{1}$,
            Xianfeng Terry Yang$\IEEEauthorrefmark{1}$
 }
	}

\maketitle
\maketitle

\begin{abstract}
Existing Advanced Driver-Assistance Systems primarily focus on the vehicle directly ahead, often overlooking potential risks from following vehicles. This oversight can lead to ineffective handling of high-risk situations, such as high-speed, closely-spaced, multi-vehicle scenarios where emergency braking by one vehicle might trigger a pile-up collision. To overcome these limitations, this study introduces a novel deep reinforcement learning-based algorithm for longitudinal control and collision avoidance. This proposed algorithm effectively considers the behavior of both leading and following vehicles. Its implementation in simulated high-risk scenarios, which involve emergency braking in dense traffic where traditional systems typically fail, has demonstrated the algorithm’s ability to prevent potential pile-up collisions, including those involving heavy-duty vehicles.x
\end{abstract}


\section{Introduction}

Advanced Driver-Assistance Systems (ADASs) are technologies designed to enhance traffic safety by monitoring the vehicle's surrounding environment and automatically intervening to prevent potential crashes should the drivers fail to respond appropriately~\cite{galvani2019history,kukkala2018advanced}. ADAS emerged in the 1970s with the adoption of the anti-lock braking system~\cite{burton1997evaluation}. Over the years, the progression of technology has enabled ADAS to expand its functionalities, including automatic emergency braking (AEB)~\cite{yang2022systematic}, stability control~\cite{ferguson2007effectiveness}, blind-spot detection\cite{liu2017radar,sotelo2008blind}, lane departure warnings~\cite{kozak2006evaluation}, adaptive cruise control (ACC)~\cite{vahidi2003research}, and traction control~\cite{de1999dynamic}. These developments have significantly improved traffic safety.


Among widely-adopted ADAS technologies, most current ACC and AEB primarily focus on the vehicle directly ahead. They often fail to consider the potential safety risks posed by vehicles following from behind~\cite{nidamanuri2021progressive}. Thus, existing ACC and AEB systems are ineffective in managing certain high-risk situations. For example, in cases where multiple vehicles travel at high speeds and are closely spaced, those behind may fail to stop safely when a vehicle in the middle of the chain suddenly activates its AEB to avoid colliding with the leading vehicle, resulting in a pile-up collision. The severity of such collisions can escalate further if one of the vehicles involved is a heavy-duty vehicle.

To mitigate collision with both leading and following vehicles, this study proposes a novel longitudinal control and collision avoidance algorithm that integrates adaptive cursing and emergency braking. Leveraging deep reinforcement learning, this approach accounts for the behavior of both preceding and following vehicles. Our methodology outperforms existing ACC and AEB systems~\cite{FR2023} in complex and hazardous driving conditions.



The main contributions of this work are:
\begin{itemize}
    \item The development of a vehicle brake and acceleration policy that enhances safety by addressing the potential safety risks from the following vehicles through the exploration of edge case collision scenarios.
    \item The development of a universally applicable algorithm designed to mitigate the incidence of serious pile-up collisions.
    \item Simulation studies show that our Deep Deterministic Policy Gradient (DDPG)-based algorithm effectively reduced collisions that traditional methods cannot avoid.
\end{itemize}

\section{Related Work}

\subsection{Adaptive Cruise Control and Automatic Emergency Braking}
Existing ACC models aim to optimize fuel consumption and reduce emissions through predictive control strategies, ensuring a comfortable ride by smoothly managing acceleration and deceleration to maintain safe distances~\cite{lu2019energy}. Their adaptability allows for effective performance under various driving conditions, incorporating real-time sensor data to seamlessly adjust to both local and highway environments~\cite{yu2022researches}. Similarly, AEB systems employ sensors and algorithms to detect imminent collisions, automatically applying brakes to either avoid or mitigate the impact severity. These systems are designed for precision, minimizing false alarms while ensuring prompt action when necessary, and are adaptable to diverse driving environments, thereby enhancing protection for vehicle occupants and pedestrians alike across different scenarios~\cite{fildes2015effectiveness}.

The integration of machine learning and artificial intelligence into the development of ACC and AEB technologies has been a crucial factor in the transition from passive to active and predictive safety systems~\cite{moujahid2018machine}. These advancements have not only enhanced the functionality and reliability of these systems but also opened new avenues for innovation in traffic safety~\cite{kim2017prediction}.

\subsection{Reinforcement learning application in ADAS}
Reinforcement Learning (RL) offers a promising approach to developing ADAS, enabling algorithms to learn optimal actions through trial and error based on feedback from the environment~\cite{wang2022deep}. Recent advancements in computational power and the development of sophisticated simulation platforms have driven significant progress in RL techniques~\cite{dosovitskiy2017carla}. Deep Reinforcement Learning (DRL) algorithms, which integrate deep neural networks with RL principles, such as Deep Q-Networks (DQN) and Deep Deterministic Policy Gradients (DDPG), have demonstrated their capability to manage the complexities of ADAS algorithms, significantly enhancing the systems' ability to navigate hazardous conditions effectively~\cite{8441758}.

The application of DRL in ADAS has been explored in various functions, including Adaptive Cruise Control (ACC)~\cite{desjardins2011cooperative}, lane-keeping assistance~\cite{sallab2016end}, and Automatic Emergency Braking (AEB)~\cite{fu2020decision,chae2017autonomous}. These studies have highlighted DRL's ability to adapt to the unpredictable behavior of other vehicles, pedestrians, and varying road conditions, thereby fostering safer and more reliable driving policies~\cite{chen2023using}. Furthermore, the integration of sensor fusion techniques with DRL has facilitated a more comprehensive understanding of the vehicle's surroundings~\cite{9106866}, enhancing the accuracy of predictions and decisions made by ADAS.

\section{Method}
In this study, a novel RL-based algorithm for longitudinal control and collision avoidance is developed to effectively manage high-risk driving scenarios. The DDPG is selected as the deep reinforcement learning model. DDPG is an algorithm that simultaneously learns a Q-function and a policy using off-policy data and an actor-critic neural network architecture. It utilizes the Bellman equation to update the Q-function, which, in turn, guides the policy learning~\cite{lillicrap2015continuous}. This model adeptly navigates complex vehicle-following situations and is capable of accommodating various vehicle types with different acceleration policies.

\begin{figure*}[!]
\centerline{\includegraphics[width=1\textwidth]{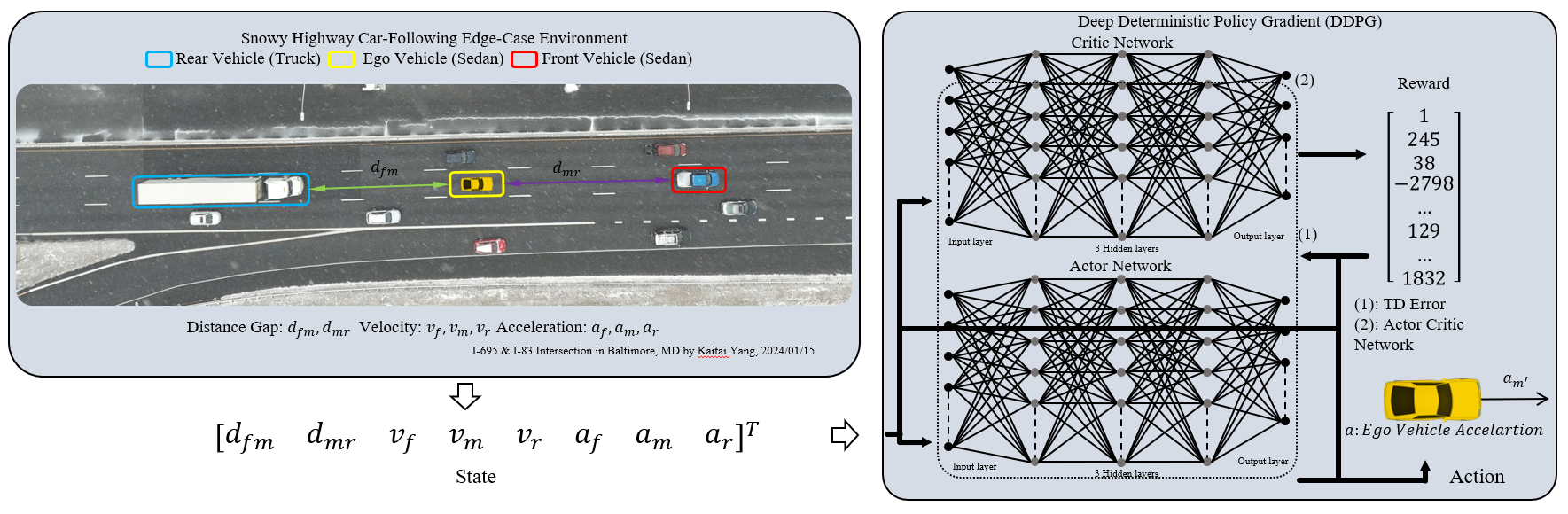}}
\caption{Model framework}
\label{CV}
\end{figure*}

\subsection{Markov decision process}
The ego vehicle RL-agent used an environment state with 8 parameters, an action, and a well-designed reward function. The car following scenario can be represented by a Markov Decision Process (MDP) with the tuple $(S,A,P_a,r_a)$
\subsubsection{S}
A set of states. $S$ represents each time step's environment state variables when we consider the RL vehicle agent as the middle ego vehicle. $S$ includes the distance between the leading vehicle and ego vehicle $d_{fm}$, distance between the ego vehicle and following vehicle $d_{mr}$, leading vehicle velocity $v_f$, ego vehicle velocity $v_m$, following vehicle velocity $v_r$, leading vehicle acceleration $a_f$, ego vehicle acceleration $a_m$, following vehicle acceleration $a_r$. The ego vehicle's sensor is only able to capture the current position of itself and the distance between itself and the leading/following vehicles. A Kalman Filter is used to estimate the velocity and acceleration.
\subsubsection{A} 
A set of actions. $A$ represent the next time step's acceleration $a_{m'}$ of the ego vehicle in current episode. 
\subsubsection{P}
The transition probability. $P_a(s,s') = Pr(s_{t+1}=s'|s_t=s,a_t=a)$, $P_a(s,s')$ is the probability of changing from state $s$ to next time step state $s'$ when take the ego vehicle acceleration action $a_{m'}$.
\subsubsection{r}
A set of rewards. $r$ represent the reward function $r(s_{t+1},s_t,a_t)$. $r$ is the expected immediate reward of taking a specific action $a$ from state $s$ to state $s'$. The  reward at a time step is 15 if there is no collision and -3000 if the vehicle collides with other vehicles. The reward is design in such way to avoid collisions.

The objective is to identify the policy function $\pi(s)$ that is able to generate the optimal action for a particular state to  maximize the expectation of cumulative future rewards:
\begin{equation}
    E\left[\sum_{t=0}^{\infty} \gamma^t r\left(s_t, s_{t+1},a_t\right)\right]
\end{equation}
where $\gamma$ is the discount factor and the range of $\gamma$ is $[0,1]$. The larger $\gamma$  motivates the RL agent to favor taking actions indefinitely, rather than postponing them early. The pseudo-code of the RL framework is shown in Algorithm~\ref{DDPGalgorithm}


\begin{algorithm}[!t]
	\caption{Proposed RL-based ADAS Algorithm} 
	\label{DDPGalgorithm} 
	\begin{algorithmic}[1]
        \REQUIRE Current time step preceding front vehicle sensor detection result $data_f$ and $ data_r$, Ego vehicle position $(x_{m},y_{m})$, Ego vehicle velocity $v_m$
        \STATE Start episode
        \FOR{$i = 1$ \TO $Maximum$ $Episode$}
        \FOR{$n = 1$ \TO $Episode$ $Steps$}
        \STATE Acquire the middle RL vehicle's position $(x_m,y_m)$, velocity $v_m$ and the approximate gap distance $d_{fm}$ $d_{mr}$ between front/rear vehicle and ego middle vehicle 
        \STATE Estimate next time step related vehicle position $(x_f,y_f)$, $(x_r,y_r)$, velocity $v_f$, $v_r$, acceleration $a_f$, $a_r$
        
        \IF{Collision happened}
        \STATE Return collision reward: $-3000$
        \STATE End current training episode
        \ELSE
        \STATE Return no collision reward: $15$
        \STATE Interact with the environment. Turn to the current episode's next time step. Return to $Step (3)$
        \ENDIF

        \ENDFOR
        \ENDFOR

	\end{algorithmic} 
\end{algorithm}
    

\section{Experiment}
\subsection{Baseline ADAS model}
A baseline ADAS model combining AEB and ACC is developed to evaluate the proposed algorithm. The Time-To-Collision (TTC) is used as the AEB and ACC engagement factor. The vehicles' velocities are kept at the same to reduce the perturbations~\cite{8884686}. Suppose the TTC between the ego vehicle and the leading vehicle is less than $1.4s$ (The Federal Highway Administration Advanced Driving Assistant System guide required)~\cite{donnell2009speed}. In that case, the ego vehicle will activate the emergency brake system. The algorithm of the baseline ADAS model is shown in Algorithm~\ref{ADASalgorithm}.
\begin{algorithm}[!]
	\caption{Baseline ADAS Algorithm} 
	\label{ADASalgorithm} 
	\begin{algorithmic}[1]
        \REQUIRE Leading vehicle related position $(x_f,y_f)$, ego  vehicle position $(x_{m},y_{m})$, leading vehicle initial velocity $v_{fi}$, ego vehicle velocity $v_{m}$, time step $\Delta t$
		\STATE Initialize Variables
        \STATE Start Episode
        \FOR{$i=1$ \TO $Maximum$ $Episode$}
        \FOR{$n=1$ \TO $Episode$ $Steps$}
        \STATE Predict preceding vehicle velocity $v_{f}$ by Kalman Filter
        \STATE Predict preceding vehicle new position $(x_{fnew},y_{fnew}) = (x_{f},y_{f}) + v_{f} * \Delta t$
        \IF{$TTC_{front\And middle}<1.4s$}
        \STATE Active ego vehicle deceleration $a_{ego}=-7.5m/s^{2}$~\cite{donnell2009speed}
        \ENDIF
        \ENDFOR
        \ENDFOR
		
	\end{algorithmic} 
\end{algorithm}

\subsection{RL Training}
The training of RL models includes two stages: exploration and exploitation. Various scenarios were employed for each stage during model training:
\begin{itemize}
    \item During the exploration stage, both the leading and following vehicles maintain their velocities based on a Gaussian Distribution. Meanwhile, the ego vehicle's velocity is determined by the acceleration output from the actor-critic neural network, and the deceleration of the following vehicles is also normally distributed. It is crucial to ensure that the neural network can learn from end-collision scenarios. Therefore, a significant portion of the training is aimed at exploring collision scenarios. This approach enables the RL agent to maximize its reward and enhance vehicle safety during the training process.
    \item In the exploitation stage, the actor-critic network guides the agent to select policies associated with higher expected cumulative future rewards. Additionally, the application of normal distribution noise to actions, with a decay factor of 0.9995, facilitates the neural network's convergence to the optimal.
\end{itemize}

\subsection{Design of scenarios}
\begin{figure}[!t]
    \centering
    \begin{subfigure}[b]{0.472\linewidth}
        \centering
        \includegraphics[width=\linewidth]{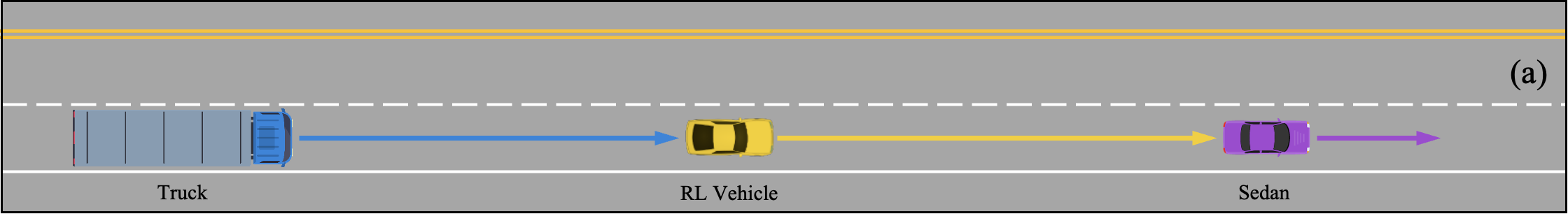}
        \caption{Highway emergency brake scenario 1}
    \end{subfigure}
    \quad
    \begin{subfigure}[b]{0.472\linewidth}
        \centering
        \includegraphics[width=\linewidth]{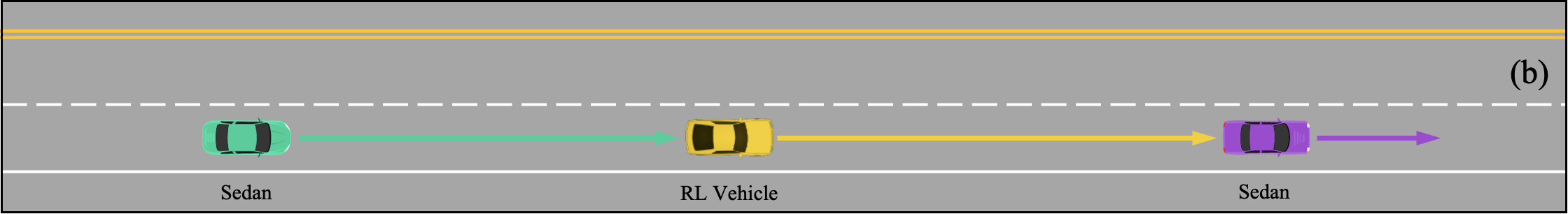}
        \caption{Highway emergency brake scenario 2}
    \end{subfigure}
    \quad
    \begin{subfigure}[b]{0.472\linewidth}
        \centering
        \includegraphics[width=\linewidth]{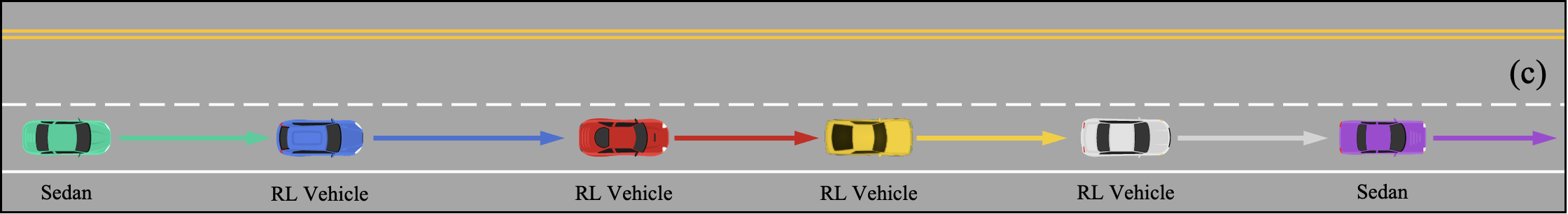}
        \caption{Highway multiple RL vehicle following scenario}
    \end{subfigure} 
    \caption{Potential dangerous collision scenarios}
    \label{SC}
\end{figure}

It is easy to train the RL algorithm in normal driving scenarios, but as mentioned before, the ADAS should be capable of handing all driving situations including high-risk ones. The study designed several scenarios aimed at causing the collision not only with the leading vehicle but also with the following vehicle to assess the algorithm's capability to address the high-risk conditions that the current baseline ADAS algorithms cannot manage, generated by digital twin environments~\cite{10619056}. The study also examines an edge case involving the following vehicle, which is heavy (compared to baseline light vehicle) and thus more prone to causing collisions with the ego vehicle. These scenarios include emergency brake scenarios (scenarios 1 and 2) and multiple vehicle following scenario (scenarios 3). A visual representation of these scenarios is provided in Fig.~\ref{SC}.

In these scenarios, vehicles are assumed to travel along a highway at a speed of 90 km/h. The first leading vehicle activates emergency braking at the 100th time step with a deceleration of $-3m/s^2$, similar to standard AEB testing scenarios~\cite{FR2023}. In Scenario 1, the blue following vehicle is configured as a heavy vehicle, while in Scenario 2, it is a light vehicle. The length of the light vehicle and heavy vehicles are assumed to be 2m and 15m, respectively, and the initial spacing between vehicles is 16m. The heavy vehicle exhibits a lower maximum deceleration with $-6m/s^2$ compared to the light vehicle with a standard AEB deceleration of $-7.5m/s^2$, In Scenario 3, all vehicles except for the last following heavy vehicle are configured as light vehicles. 

\begin{table}[!t]
    \abovetopsep = 0pt
    \aboverulesep = 0pt
    \belowrulesep = 0pt
    \belowbottomsep = 0pt
    \centering
    \begin{tabular}{c|c|c}
        \toprule
         Scenario & Key point & Vehicle Size\\
         \midrule
         \multirow{1}*{Emergency}& \multirow{1}*{Leading vehicle } & Leading vehicle: Light\\
          brake scenario 1 & emergency brake & Following vehicle: Heavy\\
         \hline
         \multirow{1}*{Emergency }& \multirow{1}*{Leading vehicle } & Leading vehicle: Light\\
         brake scenario 2 & emergency brake & Following vehicle: Light\\
         \hline
         \multirow{3}*{\shortstack{Multiple RL \\vehicles following \\scenario }}& \multirow{3}*{RL agent capability} & \multirow{3}*{\shortstack{Leading vehicle: Light \\ Following vehicle: Heavy}}\\
         ~ & ~ & ~\\
         ~ & ~ & ~\\
         \hline

    \end{tabular}
    \caption{}
    \label{dangerousscenario}
\end{table}

\subsection{Implementation}

In the baseline scenarios, all vehicles are controlled by the baseline ADAS model. Conversely, in scenarios where the proposed RL algorithm is implemented, RL agents control all vehicles except for the first leading and the last following one. The RL-controlled vehicles are trained as single agents within a scenario that involves random vehicle following. 


The proposed algorithm is implemented and evaluated in a simulation powered by the Python 3.8.10, Stable-baseline3 2.1.0, and Gym 0.26.2. The vehicle RL environment was developed based on the Gym environment "Pendulum-v1".
\paragraph{Scenario Parameters}
The vehicle positions are $(X, Y)$ and $Y=0m$, the initial positions $X_f \sim N(36,0.5)m$, $X_m \sim N(18,0.5)m$, and $X_r \sim N(0,0.5)m$, the brake deceleration of the leading vehicle is $a_{f} \sim N(-3,0.2)m/s^2$, the leading vehicle stop starts at $t_f \sim U(1,1.5)s$. The acceleration for each time step of both the leading and flowing vehicles follows a normal distribution $a \sim N(0,0.01)m/s^2$. These normally distributed parameters facilitate the ego vehicle agent's exploration of rare scenarios.

\paragraph{DDPG architecture and hyperparameters}
As shown in Fig.~\ref{CV}, four fully connected networks are established as the actor-critic neural network. Each network has three hidden layers and uses the array $[State \, dimension, 256,256,256, Action\, dimension]$ as the network node. The hyperparameters of the DDPG algorithm are shown in Tab.~\ref{hyperparameter}.

\begin{table}
    \abovetopsep = 0pt
    \aboverulesep = 0pt
    \belowrulesep = 0pt
    \belowbottomsep = 0pt
    \centering
    \begin{tabular}{c|c }
        \toprule
        Hyperparameter & Value \\
        \midrule
        \multirow{2}*{Neural Network Size} & $[State \, dim, 256,$ \\
        ~ & $256,256,Action\,dim]$\\
        \hline
        Hidden Layers Number & $3$ \\
        \hline
        Soft Update factor & $0.005$ \\
        \hline
        Memory Capacity & $10000$ \\
        \hline
        Replay Buffer Batch Size & $512$ \\
        \hline
        Discount Factor & $0.99999$ \\
        \hline
        Actor Network Learning Rate & $0.001$ \\
        \hline
        Critic Network Learning Rate & $0.002$ \\
        \hline

    \end{tabular}
    \caption{Hyperparameters for DDPG }
    \label{hyperparameter}
\end{table}

\section{Result}
Initially, the proposed RL-based algorithm was trained, followed by the implementation of the baseline ADAS models for the high-risk scenarios, as detailed in Table~\ref{dangerousscenario}. Subsequently, this algorithm was tested to assess its effectiveness.
\subsection{DDPG training result}
\begin{figure}[!t]
        \centerline{\includegraphics[width=0.5\textwidth]{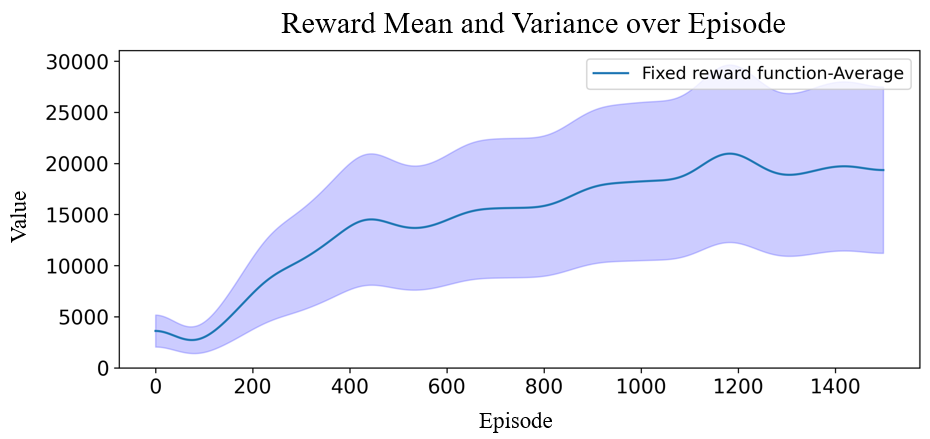}}
        \caption{Change of the Reward during RL training}
        \label{fixedreward}
    \end{figure}

The DDPG agent was trained within the designed environment. Initially, the number of the training episodes was set to 10,000 to evaluate whether the neural network was converging or overfitting, and to determine if the RL agent had successfully learned a policy capable of avoiding collisions in high-risk scenarios. The robustness of the training process was assessed using 30 different random seeds. Due to the randomness, the convergence of the DDPG model slightly differed. The dark line and light shading in the Fig.~\ref{fixedreward}. represent the mean reward and its standard deviation to $1500^{th}$ episode, respectively. The DDPG model converged at approximately the $400^{th}$ episode, with a reward of $22500$ if the vehicle successfully avoided any collisions as shown in Fig.~\ref{fixedreward}.

\subsection{Baseline ADAS simulation model result}
\begin{figure*}[!]
    \centering
    \begin{subfigure}[b]{0.3\textwidth}
        \centering
        \includegraphics[width=\linewidth]{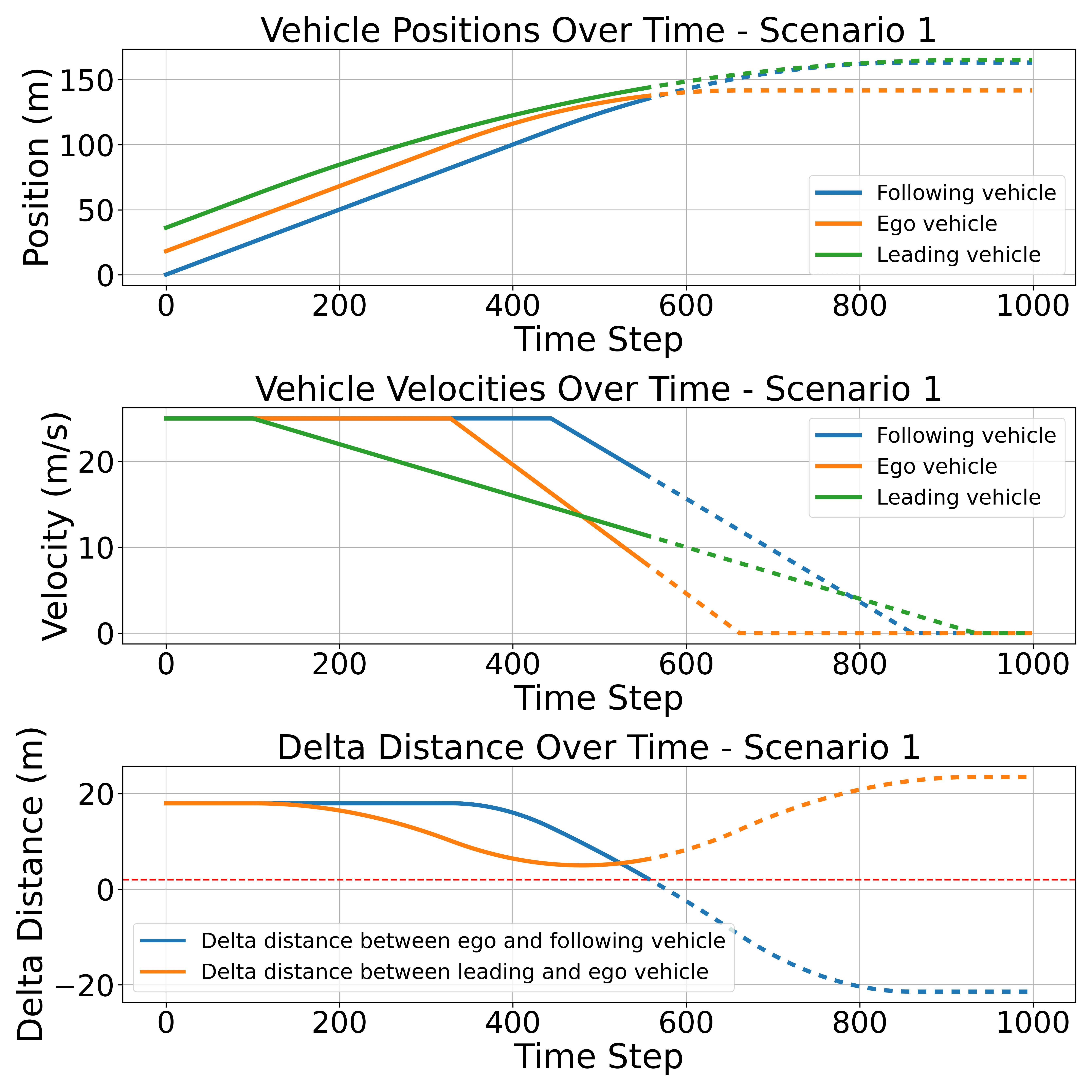}
        \caption{Highway emergency brake scenario 1}
    \end{subfigure}
    \quad
        \begin{subfigure}[b]{0.3\textwidth}
        \centering
        \includegraphics[width=\linewidth]{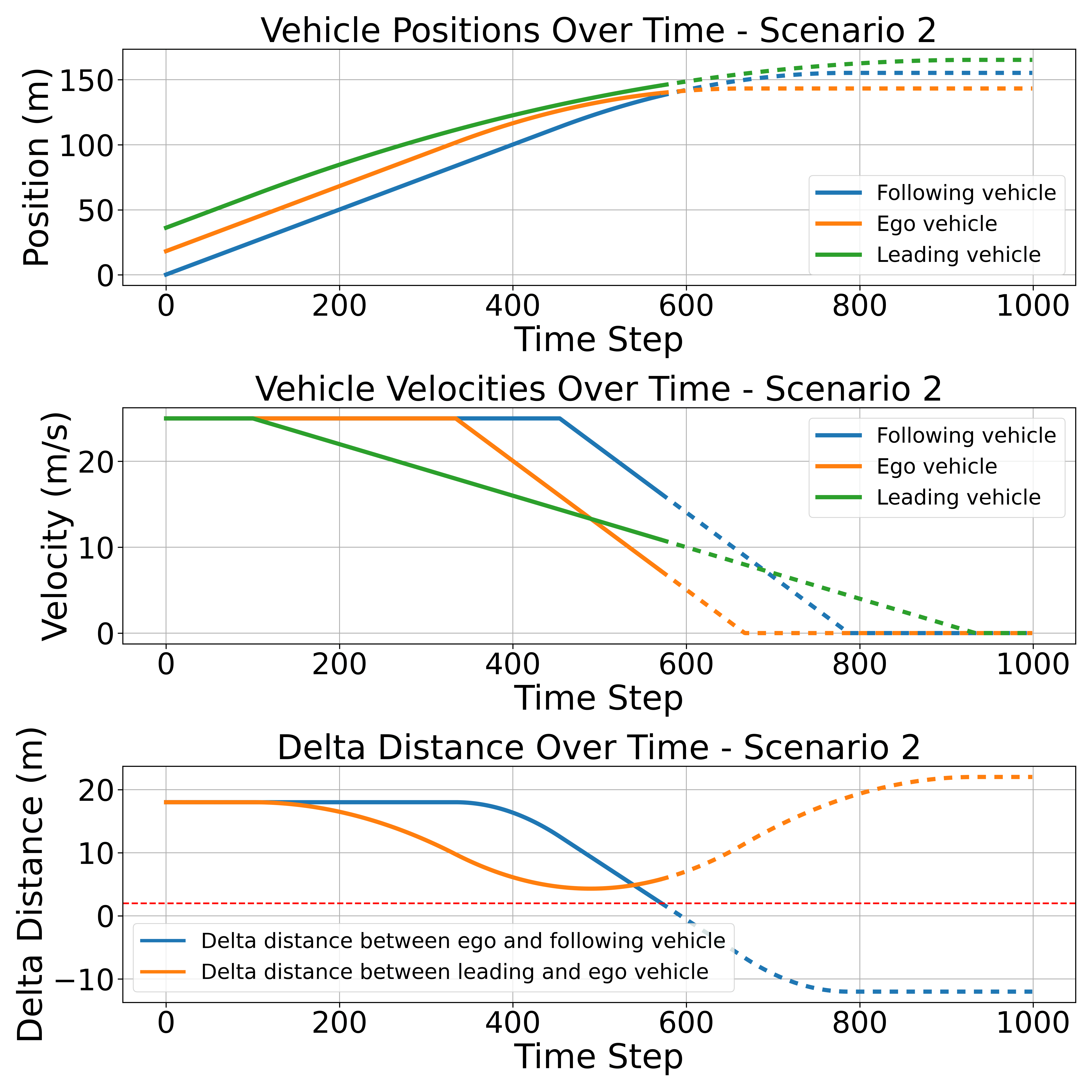}
        \caption{Highway emergency brake scenario 2}
    \end{subfigure}
    \begin{subfigure}[b]{0.3\textwidth}
        \centering
        \includegraphics[width=\linewidth]{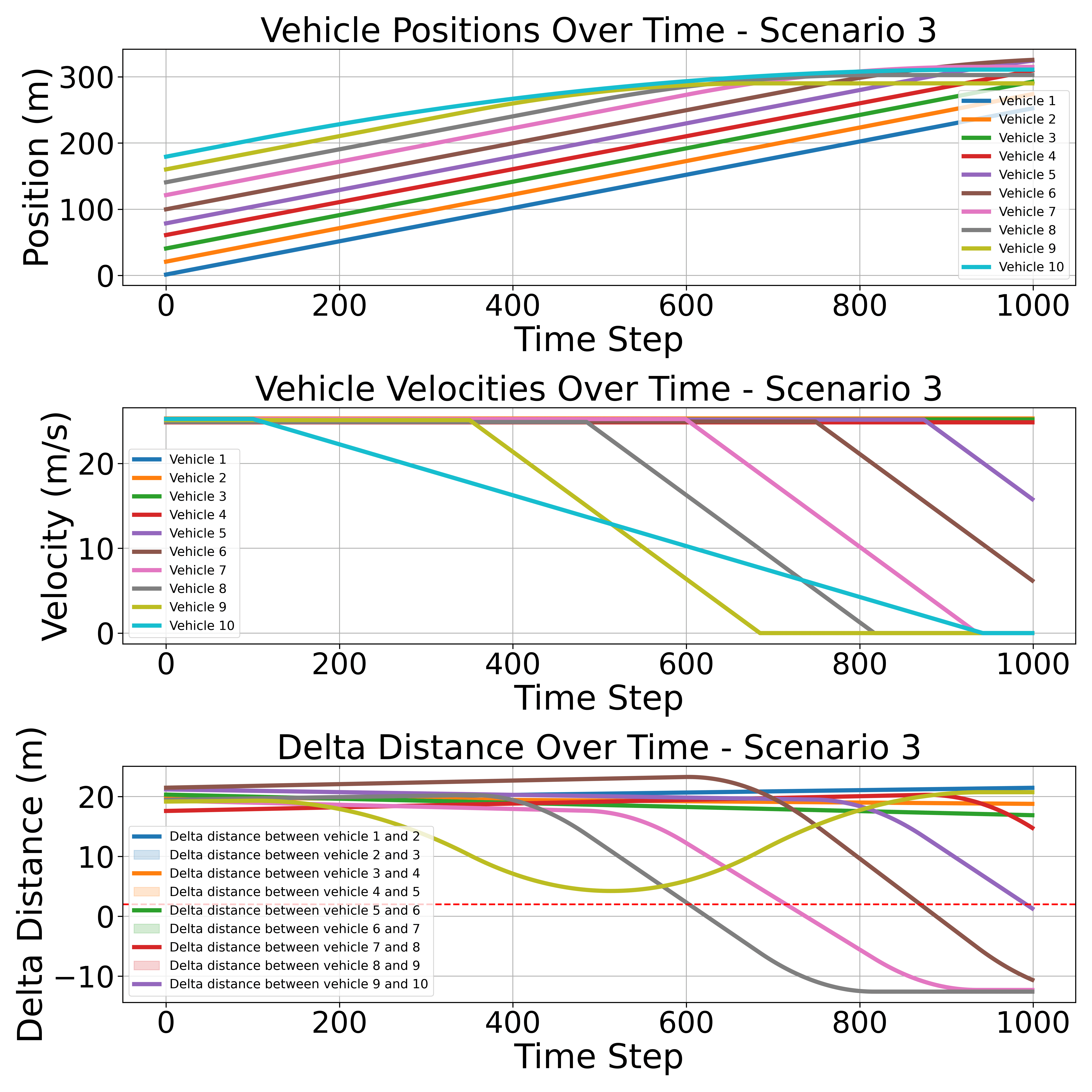}
        \caption{Highway multi-vehicles emergency brake scenario 3}
    \end{subfigure}
    \caption{Baseline ADAS algorithm implementation in proposed edge case scenarios}
    \label{Vehicleposition}
\end{figure*}

In these scenarios, all vehicles except the first leading vehicle, which activates emergency braking at a pre-defined time step with a deceleration of $-3m/s^2$, adhere to the TTC collision threshold and will initiate emergency braking at maximum deceleration. The baseline ADAS algorithm was implemented in all three scenarios. The time-space diagram, time-speed diagram, and changes in spacing between vehicles when the baseline algorithm was implemented are illustrated in Fig.~\ref{Vehicleposition}. It is evident that in both scenarios 1 and 2, the baseline ADAS can help the ego vehicle in avoiding a collision with the leading vehicle. However, it fails to prevent a collision with the following vehicle. Similarly, the failure of the baseline ADAS leads to a pile-up collision in scenario 3. One possible reason for this failure is its inability to dynamically adjust deceleration to accommodate the behavior of the following vehicle, despite sufficient space existing for all three vehicles to stop safely (as indicated by the dashed lines in  Fig.~\ref{Vehicleposition}).

\subsection{Proposed RL algorithm simulation model result}
\begin{figure*}
    \centering
    \begin{subfigure}[b]{0.3\textwidth}
        \centering
        \includegraphics[width=\linewidth]{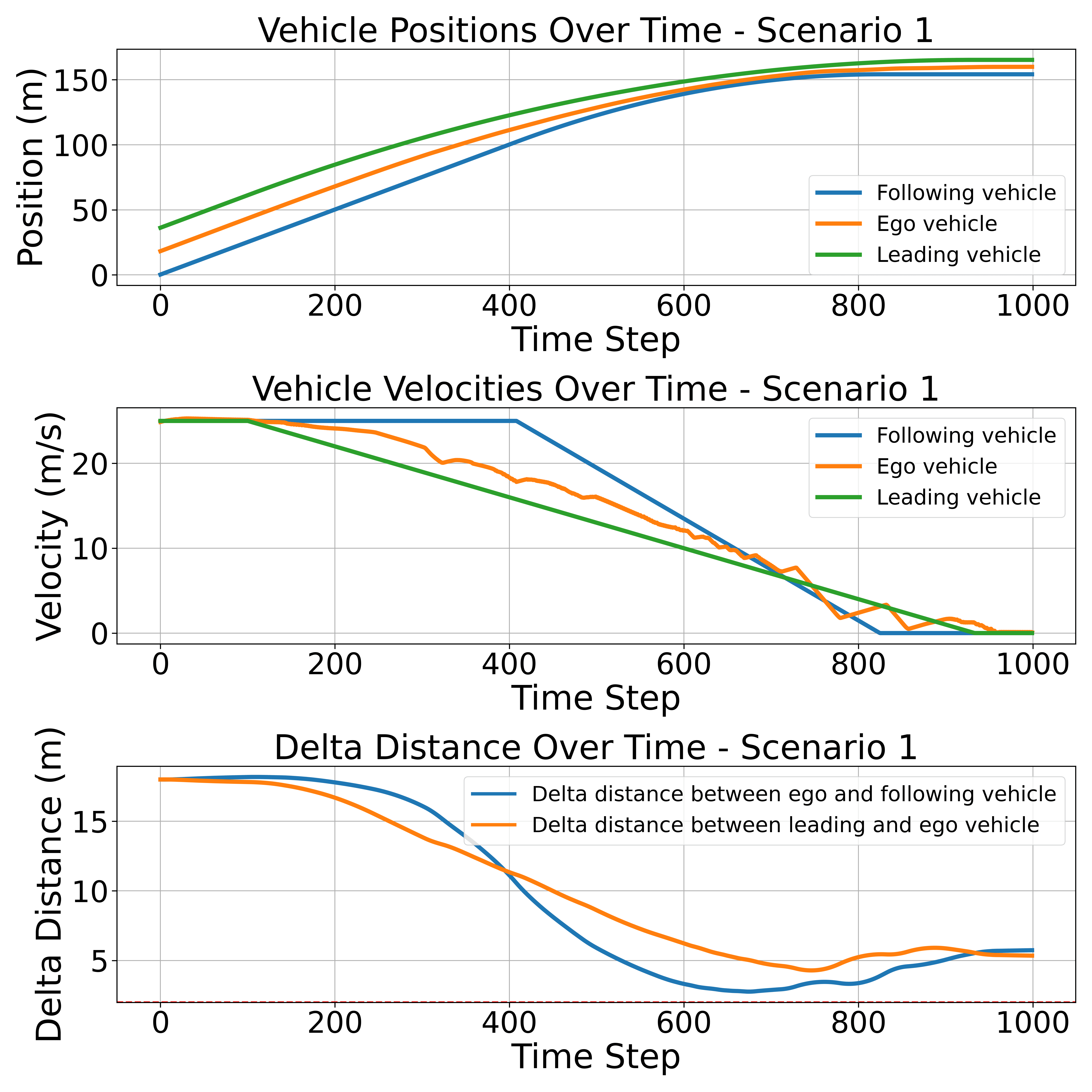}
        \caption{Highway emergency brake scenario 1}
    \end{subfigure}
    \quad  
    \begin{subfigure}[b]{0.3\textwidth}
        \centering
        \includegraphics[width=\linewidth]{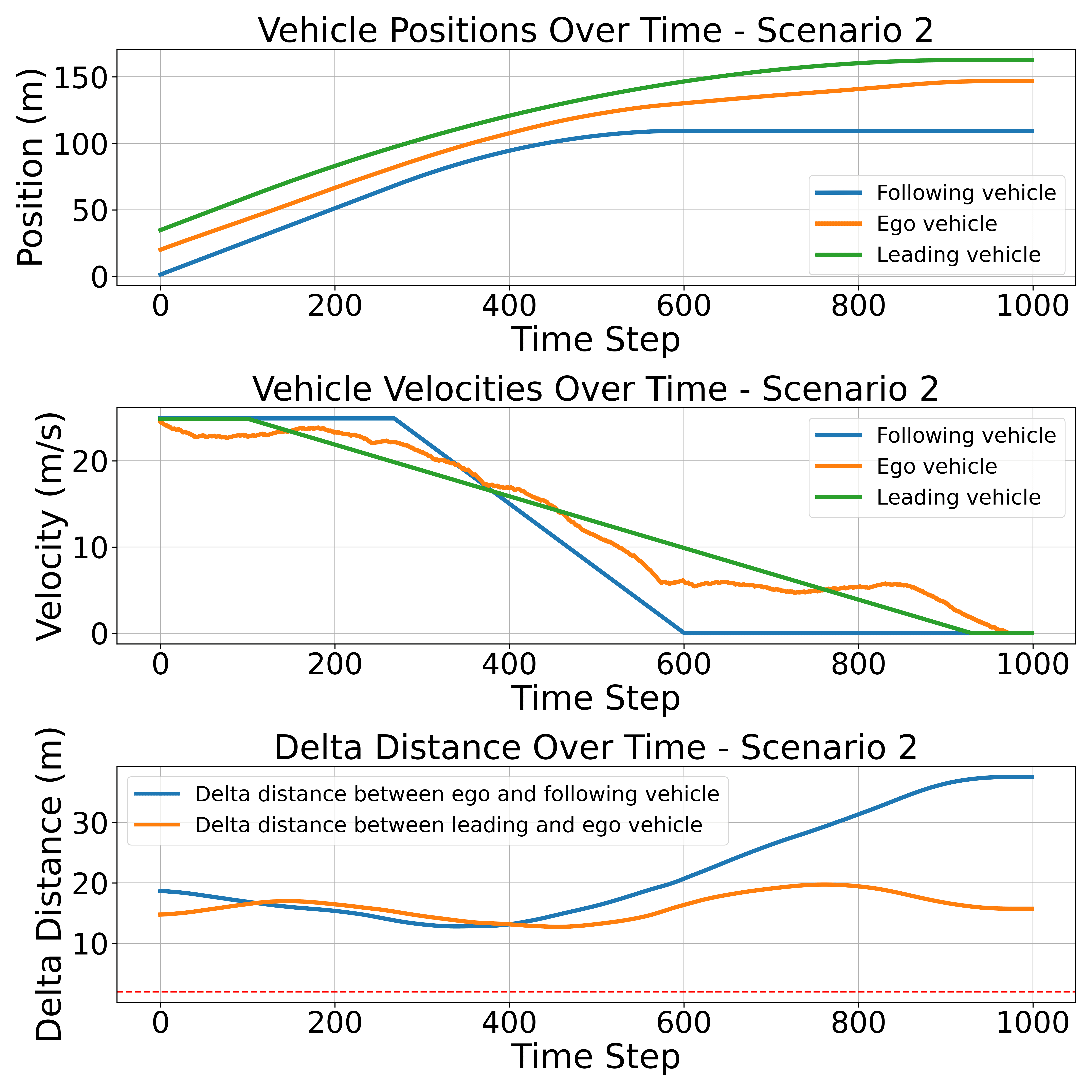}
        \caption{Highway emergency brake scenario 2}
    \end{subfigure}
    \quad  
    \begin{subfigure}[b]{0.3\textwidth}
        \centering
        \includegraphics[width=\linewidth]{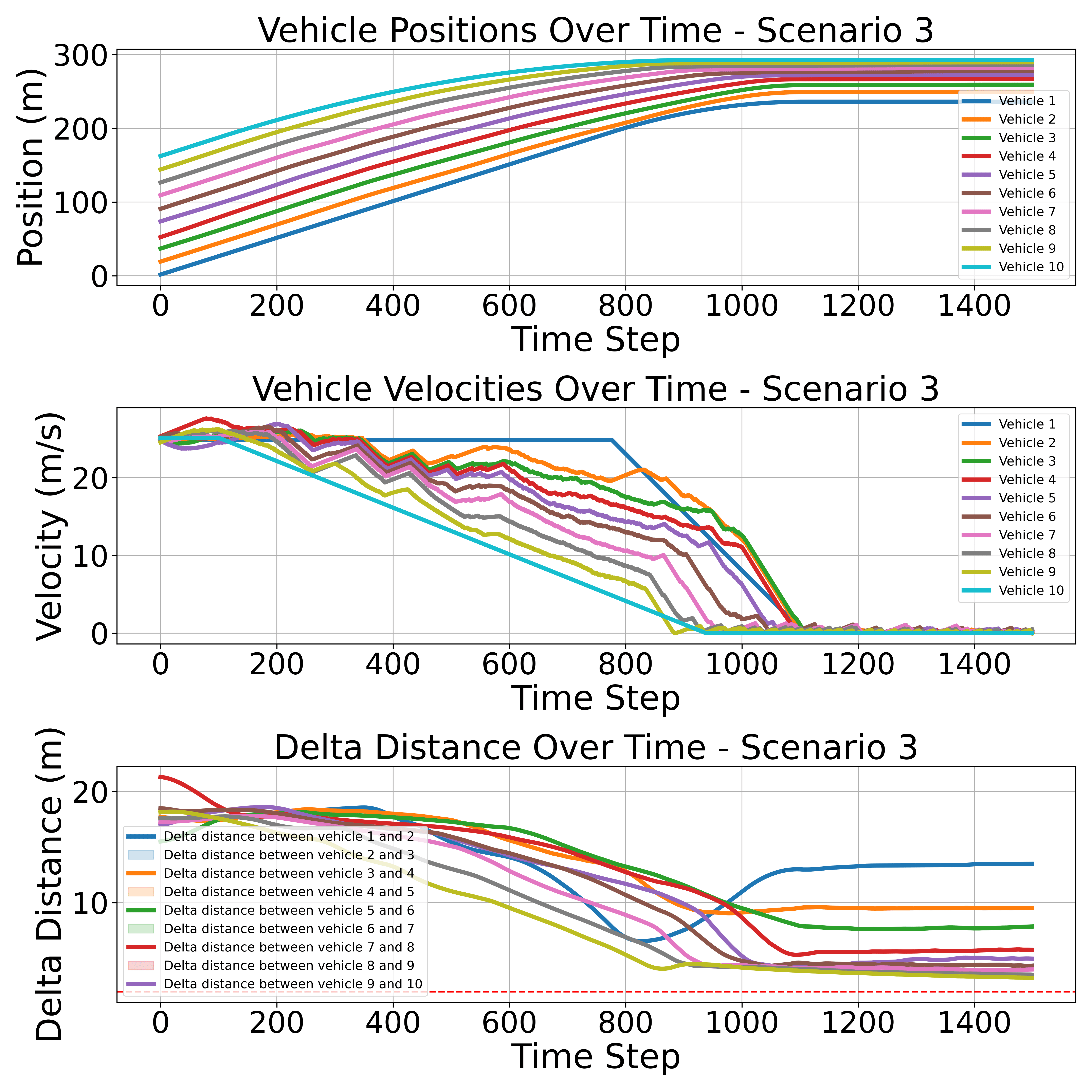}
        \caption{Highway multi-vehicles emergency brake scenario 3}
    \end{subfigure}
    \caption{Proposed DRL algorithm implementation in proposed edge case scenarios}
    \label{DRLscenario34}
\end{figure*}

Similarly, Fig.~\ref{DRLscenario34} presents diagrams when the well-trained RL algorithm was implemented in all three scenarios. In contrast to the baseline ADAS algorithm,  which relies on a fixed deceleration activated solely based on the TTC with the leading vehicle, the proposed algorithm has the capability to dynamically select different deceleration in response to the behavior of both leading and following vehicles.

\subsubsection{Emergency brake scenarios}
Unlike the baseline ADAS driving algorithm, which results in collisions in both scenarios, the proposed RL algorithm successfully avoids collisions. Specifically, in Scenario 1, there consistently  exists a one-vehicle-length gap between the green leading light vehicle and the blue following heavy vehicle. The RL algorithm optimally calculates the deceleration at each time step, allowing the yellow ego vehicle to stop within this gap without any collisions. Moreover, it is noteworthy that in emergency brake scenarios, the RL algorithms initially undergoes a sharp deceleration (from the $200^{th}$ to the $400^{th}$ time step) to trigger an early activation of the following vehicle's AEB system. This differs from the baseline algorithm, which only activates AEB upon reaching the TTC threshold. Interestingly, the RL algorithm may opt for actions that diverge from typical human driving behavior. For example, in scenario 1, it may decide to accelerate in the final few time steps, even though the spacing between the ego vehicle and the leading vehicle decreases, to ensure a safe stop of the following heavy vehicle.


\subsubsection{Multiple vehicles following scenario}
    
    
The diagrams illustrate that all vehicles stopped safely without any collisions, even though they were closely spaced. This demonstrates that the proposed RL algorithm can effectively prevent pile-up collisions triggered by the emergency braking of the leading vehicle. More importantly, the vehicles in the middle, which are controlled by the proposed RL algorithm, exhibit dynamically changing responses in terms of deceleration and acceleration. This indicates that the proposed algorithm not only effectively manages the fixed deceleration typical of conventional AEB systems but also adapts to avoid collisions by responding to the complex behaviors of leading and following vehicles.

\section{Conclusion}

This study developed a novel RL-based longitudinal control and collision avoidance algorithm designed to manage high-risk driving scenarios. Utilizing the DDPG model, the proposed algorithm effectively considers the behavior of both leading and following vehicles addressing a significant gap in existing ACC and AEB systems. The implementation of the algorithm in simulated high-risk scenarios involving emergency braking in dense traffic where traditional systems often fail, demonstrated its capability to prevent potential pile-up collisions, including those with heavy vehicles. This study contributes to the ongoing evolution of ADAS technology by incorporating artificial intelligence and shifting the paradigm from reactive systems to proactive safety mechanisms.

\bibliographystyle{IEEEtran}
\bibliography{IEEEabrv,references}

\begin{thebibliography}{10}
\providecommand{\url}[1]{#1}
\csname url@samestyle\endcsname
\providecommand{\newblock}{\relax}
\providecommand{\bibinfo}[2]{#2}
\providecommand{\BIBentrySTDinterwordspacing}{\spaceskip=0pt\relax}
\providecommand{\BIBentryALTinterwordstretchfactor}{4}
\providecommand{\BIBentryALTinterwordspacing}{\spaceskip=\fontdimen2\font plus
\BIBentryALTinterwordstretchfactor\fontdimen3\font minus \fontdimen4\font\relax}
\providecommand{\BIBforeignlanguage}[2]{{%
\expandafter\ifx\csname l@#1\endcsname\relax
\typeout{** WARNING: IEEEtran.bst: No hyphenation pattern has been}%
\typeout{** loaded for the language `#1'. Using the pattern for}%
\typeout{** the default language instead.}%
\else
\language=\csname l@#1\endcsname
\fi
#2}}
\providecommand{\BIBdecl}{\relax}
\BIBdecl

\bibitem{galvani2019history}
M.~Galvani, ``History and future of driver assistance,'' \emph{IEEE Instrumentation \& Measurement Magazine}, vol.~22, no.~1, pp. 11--16, 2019.

\bibitem{kukkala2018advanced}
V.~K. Kukkala, J.~Tunnell, S.~Pasricha, and T.~Bradley, ``Advanced driver-assistance systems: A path toward autonomous vehicles,'' \emph{IEEE Consumer Electronics Magazine}, vol.~7, no.~5, pp. 18--25, 2018.

\bibitem{burton1997evaluation}
D.~Burton, A.~Delaney, S.~Newstead, D.~Logan, and B.~Fildes, ``Evaluation of anti-lock braking systems effectiveness,'' \emph{Accident Analysis and Prevention}, vol.~29, no.~6, pp. 745--757, 1997.

\bibitem{yang2022systematic}
L.~Yang, Y.~Yang, G.~Wu, X.~Zhao, S.~Fang, X.~Liao, R.~Wang, M.~Zhang \emph{et~al.}, ``A systematic review of autonomous emergency braking system: impact factor, technology, and performance evaluation,'' \emph{Journal of advanced transportation}, vol. 2022, 2022.

\bibitem{ferguson2007effectiveness}
S.~A. Ferguson, ``The effectiveness of electronic stability control in reducing real-world crashes: a literature review,'' \emph{Traffic injury prevention}, vol.~8, no.~4, pp. 329--338, 2007.

\bibitem{liu2017radar}
G.~Liu, L.~Wang, and S.~Zou, ``A radar-based blind spot detection and warning system for driver assistance,'' in \emph{2017 IEEE 2nd Advanced Information Technology, Electronic and Automation Control Conference (IAEAC)}.\hskip 1em plus 0.5em minus 0.4em\relax IEEE, 2017, pp. 2204--2208.

\bibitem{sotelo2008blind}
M.~{\'A}. Sotelo and J.~Barriga, ``Blind spot detection using vision for automotive applications,'' \emph{Journal of Zhejiang university-SCIENCE A}, vol.~9, pp. 1369--1372, 2008.

\bibitem{kozak2006evaluation}
K.~Kozak, J.~Pohl, W.~Birk, J.~Greenberg, B.~Artz, M.~Blommer, L.~Cathey, and R.~Curry, ``Evaluation of lane departure warnings for drowsy drivers,'' in \emph{Proceedings of the human factors and ergonomics society annual meeting}, vol.~50, no.~22.\hskip 1em plus 0.5em minus 0.4em\relax Sage Publications Sage CA: Los Angeles, CA, 2006, pp. 2400--2404.

\bibitem{vahidi2003research}
A.~Vahidi and A.~Eskandarian, ``Research advances in intelligent collision avoidance and adaptive cruise control,'' \emph{IEEE transactions on intelligent transportation systems}, vol.~4, no.~3, pp. 143--153, 2003.

\bibitem{de1999dynamic}
C.~C. De~Wit and P.~Tsiotras, ``Dynamic tire friction models for vehicle traction control,'' in \emph{Proceedings of the 38th IEEE conference on decision and control (Cat. no. 99CH36304)}, vol.~4.\hskip 1em plus 0.5em minus 0.4em\relax IEEE, 1999, pp. 3746--3751.

\bibitem{nidamanuri2021progressive}
J.~Nidamanuri, C.~Nibhanupudi, R.~Assfalg, and H.~Venkataraman, ``A progressive review: Emerging technologies for adas driven solutions,'' \emph{IEEE Transactions on Intelligent Vehicles}, vol.~7, no.~2, pp. 326--341, 2021.

\bibitem{FR2023}
{National Highway Traffic Safety Administration}, ``Federal motor vehicle safety standards: Automatic emergency braking systems for light vehicles,'' Federal Register, pp. 38\,632--38\,736, 6 2023, available from: ProQuest® Congressional; Accessed: 4/23/2024.

\bibitem{lu2019energy}
C.~Lu, J.~Dong, and L.~Hu, ``Energy-efficient adaptive cruise control for electric connected and autonomous vehicles,'' \emph{IEEE Intelligent Transportation Systems Magazine}, vol.~11, no.~3, pp. 42--55, 2019.

\bibitem{yu2022researches}
L.~Yu and R.~Wang, ``Researches on adaptive cruise control system: A state of the art review,'' \emph{Proceedings of the Institution of Mechanical Engineers, Part D: Journal of Automobile Engineering}, vol. 236, no. 2-3, pp. 211--240, 2022.

\bibitem{fildes2015effectiveness}
B.~Fildes, M.~Keall, N.~Bos, A.~Lie, Y.~Page, C.~Pastor, L.~Pennisi, M.~Rizzi, P.~Thomas, and C.~Tingvall, ``Effectiveness of low speed autonomous emergency braking in real-world rear-end crashes,'' \emph{Accident Analysis \& Prevention}, vol.~81, pp. 24--29, 2015.

\bibitem{moujahid2018machine}
A.~Moujahid, M.~E. Tantaoui, M.~D. Hina, A.~Soukane, A.~Ortalda, A.~ElKhadimi, and A.~Ramdane-Cherif, ``Machine learning techniques in adas: A review,'' in \emph{2018 International Conference on Advances in Computing and Communication Engineering (ICACCE)}.\hskip 1em plus 0.5em minus 0.4em\relax IEEE, 2018, pp. 235--242.

\bibitem{kim2017prediction}
I.-H. Kim, J.-H. Bong, J.~Park, and S.~Park, ``Prediction of driver’s intention of lane change by augmenting sensor information using machine learning techniques,'' \emph{Sensors}, vol.~17, no.~6, p. 1350, 2017.

\bibitem{wang2022deep}
X.~Wang, S.~Wang, X.~Liang, D.~Zhao, J.~Huang, X.~Xu, B.~Dai, and Q.~Miao, ``Deep reinforcement learning: A survey,'' \emph{IEEE Transactions on Neural Networks and Learning Systems}, 2022.

\bibitem{dosovitskiy2017carla}
A.~Dosovitskiy, G.~Ros, F.~Codevilla, A.~Lopez, and V.~Koltun, ``Carla: An open urban driving simulator,'' in \emph{Conference on robot learning}.\hskip 1em plus 0.5em minus 0.4em\relax PMLR, 2017, pp. 1--16.

\bibitem{8441758}
A.~Moujahid, M.~ElAraki~Tantaoui, M.~D. Hina, A.~Soukane, A.~Ortalda, A.~ElKhadimi, and A.~Ramdane-Cherif, ``Machine learning techniques in adas: A review,'' in \emph{2018 International Conference on Advances in Computing and Communication Engineering (ICACCE)}, 2018, pp. 235--242.

\bibitem{desjardins2011cooperative}
C.~Desjardins and B.~Chaib-Draa, ``Cooperative adaptive cruise control: A reinforcement learning approach,'' \emph{IEEE Transactions on intelligent transportation systems}, vol.~12, no.~4, pp. 1248--1260, 2011.

\bibitem{sallab2016end}
A.~E. Sallab, M.~Abdou, E.~Perot, and S.~Yogamani, ``End-to-end deep reinforcement learning for lane keeping assist,'' \emph{arXiv preprint arXiv:1612.04340}, 2016.

\bibitem{fu2020decision}
Y.~Fu, C.~Li, F.~R. Yu, T.~H. Luan, and Y.~Zhang, ``A decision-making strategy for vehicle autonomous braking in emergency via deep reinforcement learning,'' \emph{IEEE transactions on vehicular technology}, vol.~69, no.~6, pp. 5876--5888, 2020.

\bibitem{chae2017autonomous}
H.~Chae, C.~M. Kang, B.~Kim, J.~Kim, C.~C. Chung, and J.~W. Choi, ``Autonomous braking system via deep reinforcement learning,'' in \emph{2017 IEEE 20th International conference on intelligent transportation systems (ITSC)}.\hskip 1em plus 0.5em minus 0.4em\relax IEEE, 2017, pp. 1--6.

\bibitem{chen2023using}
D.~Chen, E.~Yurtsever, K.~A. Redmill, and {\"U}.~{\"O}zg{\"u}ner, ``Using collision momentum in deep reinforcement learning based adversarial pedestrian modeling,'' in \emph{2023 IEEE Intelligent Vehicles Symposium (IV)}.\hskip 1em plus 0.5em minus 0.4em\relax IEEE, 2023, pp. 1--6.

\bibitem{9106866}
T.~Zhou, M.~Chen, and J.~Zou, ``Reinforcement learning based data fusion method for multi-sensors,'' \emph{IEEE/CAA Journal of Automatica Sinica}, vol.~7, no.~6, pp. 1489--1497, 2020.

\bibitem{lillicrap2015continuous}
T.~P. Lillicrap, J.~J. Hunt, A.~Pritzel, N.~Heess, T.~Erez, Y.~Tassa, D.~Silver, and D.~Wierstra, ``Continuous control with deep reinforcement learning,'' \emph{arXiv preprint arXiv:1509.02971}, 2015.

\bibitem{8884686}
M.~Makridis, K.~Mattas, and B.~Ciuffo, ``Response time and time headway of an adaptive cruise control. an empirical characterization and potential impacts on road capacity,'' \emph{IEEE Transactions on Intelligent Transportation Systems}, vol.~21, no.~4, pp. 1677--1686, 2020.

\bibitem{donnell2009speed}
E.~T. Donnell, S.~C. Hines, K.~M. Mahoney, R.~J. Porter, H.~McGee \emph{et~al.}, ``Speed concepts: Informational guide,'' United States. Federal Highway Administration. Office of Safety, Tech. Rep., 2009.

\bibitem{10619056}
Z.~Zhang, M.~Chen, Z.~Yang, and Y.~Liu, ``Mapping wireless networks into digital reality through joint vertical and horizontal learning,'' in \emph{2024 IFIP Networking Conference (IFIP Networking)}, 2024.

\end{thebibliography}


 




\vfill

\end{document}